\documentclass[letterpaper, 10pt, conference]{ieeeconf}      % Use this line for a4 paper

\IEEEoverridecommandlockouts                             
\usepackage{xcolor,soul,framed} %,caption
\usepackage{moreverb}
\usepackage{lipsum}
\usepackage{colortbl}
\usepackage{xcolor}
\usepackage{lineno,hyperref}
\usepackage{times}
\usepackage{multicol}
\usepackage{multirow}
\usepackage{graphicx}
\usepackage{amsbsy}
\usepackage{color}
\usepackage{epsfig}
\usepackage{epstopdf}
\usepackage{adjustbox}
\usepackage{nicefrac}
\usepackage{float}
\usepackage{amsmath}
\usepackage{amsfonts}
\usepackage{mathtools}
\usepackage{makecell}
\usepackage{multirow}
\usepackage{booktabs}
\usepackage{siunitx}
\usepackage{algorithm}
\usepackage{algpseudocode}
\usepackage{rotating}
\usepackage{relsize}
\usepackage{wrapfig}
\usepackage{forest}
\usepackage{subfigure}

\usepackage{bm}
\usepackage{fancyhdr} 
\usepackage{setspace} 
\usepackage{xspace}
\usepackage{afterpage}% http://ctan.org/pkg/afterpage
\newlength{\oldtextfloatsep}

\usepackage{pifont}

\newcommand{\figref}[1]{Figure ~\ref{#1}}
\newcommand{\tabref}[1]{Table ~\ref{#1}}
\newcommand{\equatref}[1]{Eq. ~\eqref{#1}}
\usepackage{hyperref}

\usepackage{eso-pic}
\newcommand\AtPageUpperMyright[1]{\AtPageUpperLeft{%
 \put(\LenToUnit{2cm},\LenToUnit{-1cm}){%
     \parbox{\textwidth}{\centering\fontsize{9}{11}\selectfont #1}}%
 }}%
\newcommand{\conf}[1]{%
\AddToShipoutPictureBG*{%
\AtPageUpperMyright{#1}
}
}

\newcommand\alberto[1]{\textcolor{black}{#1}}

\title{\LARGE \bf
Active Illumination for Visual Ego-Motion Estimation in the Dark
}

\author{Francesco Crocetti, Alberto Dionigi, Raffaele Brilli, Gabriele Costante, Paolo Valigi
\thanks{$^{1}$The authors are with the Department of Engineering, University of Perugia, 06125 Perugia, Italy. 
{\tt\footnotesize email:\{francesco.crocetti, alberto.dionigi, gabriele.costante, paolo.valigi\}@unipg.it; raffaele.brilli@dottorandi.unipg.it.}}
}

\begin{document}

\maketitle

\conf{This work has been accepted to the IEEE International Conference on Robotics and Automation (ICRA). This is an archival version of our paper. Please cite the published version DOI: \url{https://doi.org/10.1109/ICRA55743.2025.11127536}}

\thispagestyle{empty}
\pagestyle{empty}

%%%%%%%%%%%%%%%%%%%%%%%%%%%%%%%%%%%%%%%%%%%%%%%%%%%%%%%%%%%%%%%%%%%%%%%%%%%%%%%%
\begin{abstract}
Visual Odometry (VO) and Visual SLAM (V-SLAM) systems often struggle in low-light and dark environments due to the lack of robust visual features. In this paper, we propose a novel active illumination framework to enhance the performance of VO and V-SLAM algorithms in these challenging conditions. The developed approach dynamically controls a moving light source to illuminate highly textured areas, thereby improving feature extraction and tracking. Specifically, a detector block, which incorporates a deep learning-based enhancing network, identifies regions with relevant features. Then, a pan-tilt controller is responsible for guiding the light beam toward these areas, so that to provide information-rich images to the ego-motion estimation algorithm. Experimental results on a real robotic platform demonstrate the effectiveness of the proposed method, showing a reduction in the pose estimation error up to 75\% with respect to a traditional fixed lighting technique.

%OLD (fist submission)
%Visual Odometry (VO) and Visual SLAM (V-SLAM) systems often struggle in low-light or dark environments due to the lack of discernible visual features. This paper proposes a novel active illumination framework to improve the performances of VO and V-SLAM algorithms in such challenging conditions. The approach dynamically controls a mounted light source to illuminate texture-rich areas, thereby improving feature extraction and tracking. A deep learning-based image enhancement strategy is employed to identify regions with high feature density, and an information-driven controller guides the light beam toward these areas. Extensive experiments in real scenarios demonstrate the effectiveness of the proposed method, showcasing significant improvements in pose estimation accuracy and robustness compared to traditional fixed lighting techniques.

\end{abstract}

%%%%%%%%%%%%%%%%%%%%%%%%%%%%%%%%%%%%%%%%%%%%%%%%%%%%%%%%%%%%%%%%%%%%%%%%%%%%%%%%
\section{INTRODUCTION}
Vision-based pose estimation is one of the most widespread strategies to achieve mobile robot localization. Several effective Visual Odometry (VO) and Visual SLAM (V-SLAM) approaches have flourished in the last decades \cite{legittimo2023benchmark}, and the recent emergence of visual-inertial techniques has shown even more impressive results \cite{qin2018vins, geneva2020openvins}.

The effectiveness of VO and V-SLAM solutions depends on the capability to extract robust and highly-descriptive visual features. These can be sparse \cite{forster2014svo, ORBSLAM3_TRO}, dense \cite{newcombe2011dtam, engel2014lsd}, or even learning-based \cite{detone2018superpoint, mollica2023integrating}. Most of the state-of-the-art works assume that the operating conditions of vision-based estimation are nearly ideal, \textit{i.e., }texture-rich scenes with proper lighting conditions. However, even a slight performance drop in the feature extraction and tracking modules might cause the failure of the entire estimation pipeline.

Nonetheless, numerous applications require robotic platforms to operate in far-from-ideal conditions, such as those in disaster management or underground exploration. Low-light or completely dark environments, in particular, are the worst-case scenarios for vision-based systems since they almost completely neglect the possibility of extracting visual cues. 

A straightforward solution could be equipping the robot with a powerful wide-beam light source to illuminate the entire scene captured by the camera \cite{mansouri2020deploying}. However, this approach has two significant drawbacks: (i) it is not energy-efficient, as battery power is wasted illuminating areas without textures (\textit{e.g.}, flat surfaces), and (ii) in large environments, the light power may not be sufficient to properly illuminate the entire scene due to the excessive light scattering.  
%Unlike natural illumination, artificial light can cause reflections, excessive intensity saturation, and artifacts, leading to weak and unreliable features \cite{kasper2019benchmark}; conversely, a fixed light source with a focused narrow beam is more energy-efficient and would properly illuminate faraway portions of the scene providing better visibility. However since most of the scene remains not illuminated, feature tracking can be easily lost.
Conversely, a fixed light source with a focused narrow beam is more energy-efficient and would properly illuminate faraway portions of the scene, providing better visibility. However since most of the scene remains not illuminated, feature tracking can be easily lost. 
% On the other hand, artificial light can cause reflections, excessive intensity saturation, and artifacts, leading to weak and unreliable features \cite{kasper2019benchmark}.

 \begin{figure}[t]
     \centering
     \includegraphics[width=\linewidth]{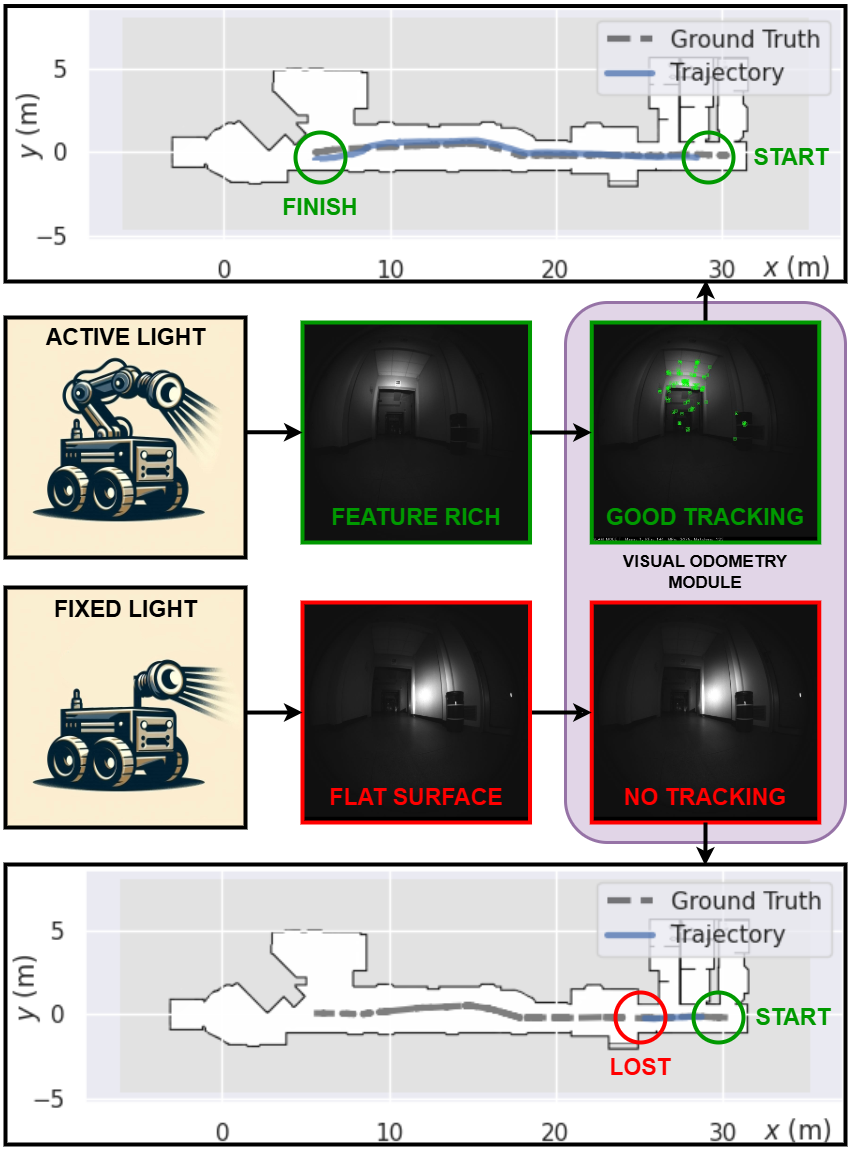}
     \caption{Navigating in dark conditions with a fixed light source may result in illuminating flat surfaces, which could cause a VO method to fail. To address this drawback, we propose the use of an active controller to guide the light beam to richly textured areas, improving the quality of the overall scene captured by the vision sensor and, as a consequence, the accuracy of the pose estimation algorithm.}
     \label{fig:intro_fig}
     \vspace{-1.5em}
\end{figure}

%Driven by the aforementioned considerations, this work proposes an alternative solution that overcomes the limitations of state-of-the-art systems. 
Motivated by the aforementioned challenges, this work introduces a new approach that advances the state-of-the-art. Specifically, as shown in Figure \ref{fig:intro_fig}, we equip the robot with a narrow-beam light source mounted on a robotic arm and develop a novel approach to identify the area in the scene expected to contain the highest amount of salient features. This allows us to define a set point for controlling the robotic arm and dynamically adjust the direction of the light beam. 
%We employ a deep learning-based image enhancement strategy to periodically obtain an illuminated image of the scene and extract features from it. Subsequently, we identify the region with the highest number of features and adjust the beam orientation accordingly via the robotic arm. 
The camera images are continuously sampled and processed at a predefined rate by a deep learning-based enhancing network. This generates a stream of enhanced images, which are then analyzed to identify the target area to focus the light beam moving a robotic arm equipped with a led spotlight end-effector. 

To the best of our knowledge, this is the first approach to achieve active light beam adaptation through a robotic arm for vision-based pose estimation.

The rest of the paper is organized as follows: in Section \ref{sec:related_works} we provide a comprehensive literature review on the problem considered.
Then, in Sections \ref{sec:methodology} we detail the proposed methodology, while
in Section \ref{sec:experiments} we outline the experimental setup and discuss the results obtained. 
Finally, in Section \ref{sec:conclusion} we summarize our findings and draw conclusions.

\section{RELATED WORKS}\label{sec:related_works}
In the context of robotic navigation, the capability of perceiving the environment by relying only on visual sensors is still one of the most open challenges, especially in complex harsh scenarios.
In the following, we provide a comprehensive literature review on V-SLAM and VO algorithms.
Then, as we propose an active approach, we also provide a literature overview of approaches based on this paradigm. 
Finally, we highlight the contribution of our work.
% , especially regarding robustness and pose accuracy. In the following, we provide a comprehensive literature review on visual SLAM and visual odometry algorithms focusing on harsh scenarios.

% {In recent years, different methods and solutions have evolved according to sensor and computation capabilities advancement.}

\textbf{Visual SLAM and VO.} 
V-SLAM and VO algorithms have been actively explored and improved from the first works \cite{davison2007monoslam} and \cite{nister2004visual}. ORB-SLAM3 \cite{ORBSLAM3_TRO} and LDSO \cite{gao2018ldso} represent milestones in the State-of-the-Art for feature-based and direct methods, respectively. Afterwards, the exploitation of depth cameras and Inertial Measurement Units (IMU) lead to Visual-Inertial (VI) methods, like MSCKF \cite{mourikis2007multi}, OKVIS \cite{leutenegger2015keyframe}, VINS-Mono \cite{qin2018vins}, and OPEN-VINS \cite{geneva2020openvins}. 
%use nonlinear filters like the Extended Kalman Filter (EKF) to fuse visual and inertial information \cite{schmidt1966application}, as well as 3D map optimization techniques \cite{mur2017visual}.
Despite the advancements in accuracy and overall robustness, harsh illumination conditions are still an open challenge \cite{bujanca2021robust}. One critical element is the dependence on ideal illumination to perceive sufficient environmental information \cite{park2017illumination}. 

Deep Learning (DL) approaches leverage Convolutional Neural Networks (CNNs) to mitigate the non-ideal conditions: they can compute features invariant to geometric and photometric changes, including illumination, background, viewpoint, and scale \cite{wang2020approaches}. These techniques are exploited by end-to-end approaches that can be easily adapted
to different setups (monocular, stereo, and RGB-D) such as DROID-SLAM \cite{teed2021droid} that leverage on recurrent iterative updates of camera pose and pixel-wise depth through a Dense Bundle Adjustment layer. Other approaches take advantage of Graph Convolutional Neural Networks (GCNs)  and RGB-D sensors \cite{derr2018signed}. Hybrid approaches like DXSLAM \cite{li2020dxslam} use DL-based methods for extracting features, which are then integrated into geometric VO/SLAM. 
However, no works consider low-light environments except for some adaptations for underwater applications \cite{zhang2022visual} that cannot be used in the context of ground robotics.

%a seminal study of the effectiveness of Hybrid approach is presented in \textcolor{blue}{[LF$^2$SLAM]}. The study proposed in \textcolor{blue}{[Lavoro ORBSLAM vs. DSO]} indicates that the combined use of different light intensities and processing images with an enhancing GAN network would dramatically impact performance in pose estimation.

\textbf{Active Approaches in low-light Conditions.}
Dealing with dark scenarios by only equipping the robotic platform with static light sources poses severe constraints on the effectiveness of V-SLAM methods, especially in challenging environments where the features are not uniformly distributed in the scene. Conversely, an active lighting approach that exploits the movement of the light source could significantly improve the performance of the vision-based algorithms. %in dark conditions. 

In general, the use of an active perception system \cite{bajcsy1988active} consists of leveraging the movement of the robot to actively acquire data from the environment to obtain information more relevant to the specific task. 
The literature presents classical and learned approaches for the design of the active controller. Classical approaches are often based on information-driven systems \cite{eidenberger2010active, stampfer2012information} to guide the visual algorithm toward more informative regions. On the other hand, learned ones mostly rely on Deep Reinforcement Learning (DRL) \cite{dionigi2022vat,dionigi2024d} to train suitable end-to-end policies that directly map input data to control actions. While the latter strategies achieved impressive performance results, they lack robustness, which is more crucial in challenging scenarios like the one considered here. Consequently, the few contributions that propose an active method in this setting are all information-driven.

The authors in \cite{wang2022automated}  propose a novel automated camera-exposure control algorithm to enhance vision-based localization in complex environments with dynamic illumination. However, dynamically changing the exposure or, in general,  sensor parameters (\textit{i.e.}, shutter speed, sensor sensitivity) could lead to a higher level of noise and/or a drop in frame rate \cite{kim2020proactive}, which can negatively impact vision algorithms. The work in \cite{kong2021direct} adopts near-infrared (NIR) light for visual SLAM in challenging lighting conditions, achieving promising results. Nevertheless, NIR images might exhibit less texture than visible light images, particularly in low-texture environments. Additionally, the requirement for specialized NIR devices and the reliance on depth sensors can limit the applicability of this approach.

% Additionally, in very low light conditions, the frame rate could drop to very low values, further affecting the performance. 
In \cite{wang2023active}, a gimbal camera is used as an active device to enhance V-SLAM accuracy and robustness in challenging environments. In particular, the authors introduce a map representation based on feature distribution-weighted Fisher information coupled with an information-gradient-based local view planner to move the camera view for obtaining maximal environmental information. Nevertheless, only environments with poor features are considered, while dark or low-light conditions are not considered.

% \textcolor{red}{Automatic explosure \cite{wang2022automated} could be considered an active approach for adapting the camera to scenarions charcterized by varing lighting conditions. However, when ... . }

% \textcolor{red}{ This article \cite{wang2023active} develops a novel view planning ap
% proach of actively perceiving areas with maximal informa
% tion to address the mentioned problem; a gimbal camera is
 % used as the main sensor. }

 \begin{figure*}[t]
     \centering
     \includegraphics[width=\linewidth]{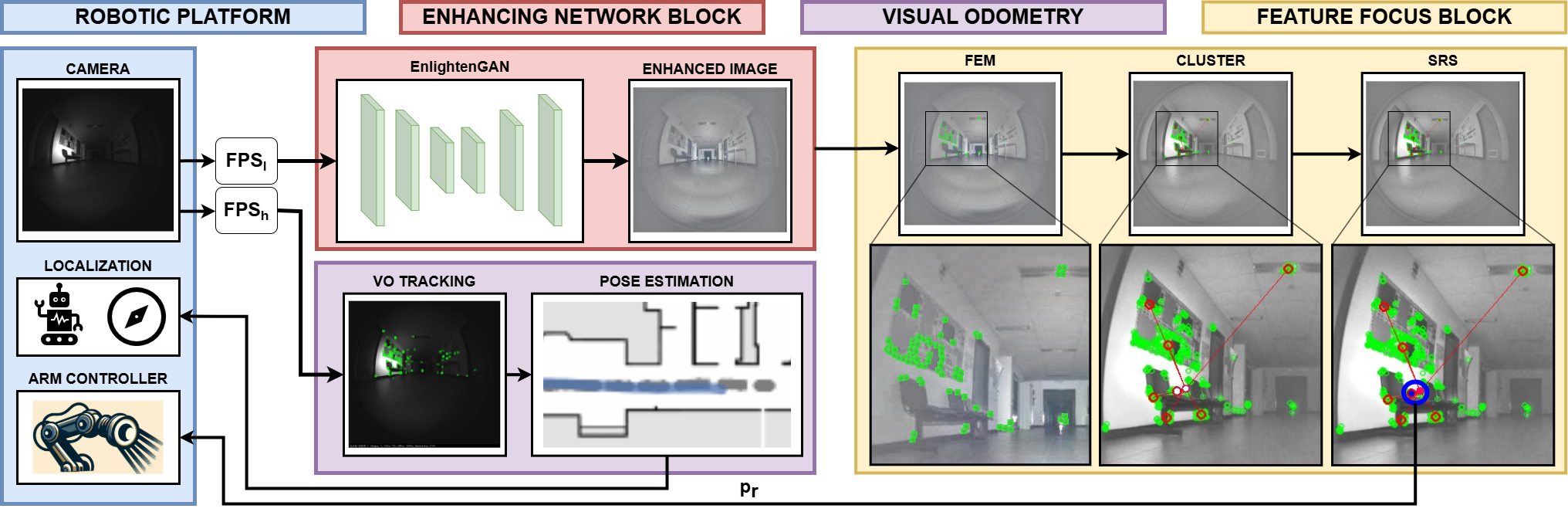}\\
     \caption{The image depicts the proposed active illumination framework, designed to enhance the performance of VO and V-SLAM algorithms in low-light environments. The framework employs two parallel image streams: a high-rate stream for real-time V-SLAM processing and a low-rate stream for feature analysis. The low-rate stream is enhanced by EnlightenGAN and then processed by the Feature Focus Block to identify areas rich in visual features. This information guides a 2-axis moving light source to dynamically illuminate these areas, ensuring the high-rate V-SLAM pipeline receives images with improved feature visibility. The adaptive illumination strategy increases the accuracy and robustness in challenging low-light conditions.}
     %This adaptive illumination strategy aims to bolster V-SLAM robustness and accuracy in challenging lighting conditions.
     \label{fig:framework_overview}
     \vspace{-1em}
\end{figure*}

\subsection{Contributions}
Although active approaches have shown significant results in enhancing the performance of visual algorithms in challenging environments, to the best of our knowledge, no previous work has proposed an active strategy specifically tailored for low-light conditions. Therefore, we introduce a new framework that incorporates a novel information-driven active approach capable of illuminating areas with high feature density. Specifically:

% \textcolor{red}{While important results have been demonstrated by active approaches for enhancing the performance of visual algorithms in challenging environments, to the best of our knowledge there are no previous works that proposed an active strategy specifically tailored for low-light conditions. Consequently, we present a new framework that incorporates a novel information-driven with a moving light source to illuminate areas with an high feature density. Specifically:}
%Thanks to the developed active methodologies, all the aforementioned works contributed with important improvements in terms of localization performance. However, since UWB-based infrastructure-less relative localization systems {\color{blue}are only recently emerging}, to the best of our knowledge there are not in literature active approaches specifically tailored for this task. Therefore, in this work we present multiple contributions:

\begin{itemize}
    \item We design a novel method to dynamically identify in low-light conditions the portion of the image that contains the highest number of features;
    \item We propose a new information-driven active method that controls a movable light source in order to illuminate texture-rich areas of the environment;
    \item We build a robotic platform equipped with a controllable light source, and through extensive real-world experiments, we demonstrate that our approach significantly outperforms the current state-of-the-art methods.
\end{itemize}

\section{METHODOLOGY}\label{sec:methodology}

\subsection{Problem Statement}
In this work, we consider scenarios characterized by total darkness, and a mobile robot equipped with a directional robotic arm that mounts the light source.
The objective of the proposed active illumination method is to provide a robotic platform with a suitable control policy for the light beam direction to obtain the best possible estimation of the robot pose from a VO or V-SLAM algorithm.  %During navigation, the robot must localize itself using only a monocular camera, and the goal is to adapt the light source direction to minimize the localization error of the visual algorithm.

To this aim, we design a novel framework \ref{subsec:framework} for visual localization in low-light conditions that leverages image-enhancing GAN networks for identifying areas with high feature density \ref{subsub:enhancing}, and a novel active lighting strategy \ref{subsec:active} coupled with an arm controller \ref{subsec:arm_controller} for illuminating such areas.

% lighting up selecting the point for positioning the robotic arm, equipped with a high power light source. 

\subsection{Active Lightning Framework}\label{subsec:framework}

The developed framework depicted in \figref{fig:framework_overview} is centered on a key element, the 2-axis moving light source. This component enables the robot to dynamically adjust the illumination of its surroundings. The active lighting system, detailed further in section \ref{subsec:exp_setup}, is a closed-loop controlled device that moves the beam light in a dark environment.
%to mitigate the lack of features the V-SLAM algorithm may encounter if the scenario is lit just by a fixed light. 
While the V-SLAM algorithm keeps tracking and identifying features and descriptors to estimate the pose, the image stream is fed into a low-rate resampler that outputs the original image but a lower FPS. The resampler is the entry point for a more comprehensive processing pipeline that runs parallel to the pose estimation process. This pipeline consists of two primary stages: an ``Enhancing Network Block (ENB)'' and the ``Feature Focus Block (FFB)'', responsible for generating the control reference signals for the 2-axis light beam device. In the following, we provide a detailed discussion of these three main components.

\subsubsection{Enhancing Network Block}\label{subsub:enhancing}
The approach used to enhance dark images, EnlightenGAN \cite{jiang2021enlightengan}, is an unsupervised framework demonstrating good generalization capabilities without paired training data; the method has been proven to be beneficial in significantly dark scenarios.
% The training process relies on four losses that can be divided into two main categories: Self Preserving Losses $L_{SFP}$ demanded to maintain the image's content and perceptual qualities and the Generator Losses $L_{G}$, to guarantee realistic image enhanced. 
% \begin{equation}
% \mathcal{L}_{\text{GAN}} = \mathcal{L}_{SFP}^{\text{Global}} + \mathcal{L}_{SFP}^{\text{Local}} + \mathcal{L}_{G}^{\text{Global}} + \mathcal{L}_{G}^{\text{Local}} \; ,
% \end{equation}
%  where $\mathcal{L}_SFP$ Losses Preserve image content by minimizing the difference in deep features (from a pre-trained VGG network \cite{simonyan2014very}) between the enhanced output and the original low-light input, at both global (whole image) and local (patch) levels and 
% $\mathcal{L}_G$ are adversarial losses that encourage the generator to produce enhanced images indistinguishable from real normal-light images. 
The results achieved on publicly available datasets and the training approach, which does not need paired images, directed us in selecting this network. Furthermore, it should be noted that although we selected EnlightenGAN, other approaches can be easily integrated in our pipeline.

EnlightenGan, coupled with the other elements Feature Focus Module, demands substantial computational resources and on hardware-constrained devices, the computational time is not compatible with the VO/V-SLAM pipelines. 
To overcome this problem, we split the image stream from the camera into two substreams: the high-rate $FPS_h$ stream directly forwarded to the VO pipeline and the low-rate stream $FPS_l$, which is processed by EnlightenGAN. The lower rate stream is computed by a resampler block whose sampling time $T_s$ is a hyperparameter: the enhanced image, computed at a lower rate, is used only to determine the area in the scene with most features (see next section \ref{subsec:active}). By directing the light to that area using the robotic arm, the high-rate image stream can effectively run the VO algorithm. This approach reduces failures, as the light is consistently focused on the region with the most features.

\begin{algorithm}[t]
\label{algo:active}
\begin{algorithmic}[1]
\State \textbf{Input:} Camera Image $\mathcal{I}$
% \State \textbf{Output:} Joint 1 Pulse Position Modulation $\mathcal{J}_1$; Joint 2 Pulse Position Modulation $\mathcal{J}_2$
\State \textbf{Output:} Target Point: p\_r
\State \textbf{Initialize:} Previous Target Point p\_last $\gets$ \text{None}; \\
Centroids $C \gets \left[\,\right]$; Features $N \gets \left[\,\right]$; Distances $D \gets \left[\,\right]$
\While{True}
    % \State $\mathcal{I}$ \{Acquire Image\}
    \State $\mathcal{I}$\_enh $\gets$ EnlightenGAN($\mathcal{I}$) \{Enhancement\}
    \State key\_points $\gets$ feature\_extractor($\mathcal{I}$\_enh) \{FEM\}
    \State $C$, $N$ $\gets$ cluster(key\_points) \{SRS\}
    \State $D \gets \left[\,\right]$ \{Reset $D$\}
    \For{$i=0$, size($C$)}
        \If{p\_last is None}
            \State $D$.append(0) \{First Iteration\}
        \Else
            \State $D$.append($||$p\_last$ - C[i]||$) \{Compute $d_i$\}
        \EndIf
    \EndFor
    \State $N_{norm} \gets ||N||$; $D_{norm} \gets ||D||$ \{Eq. \eqref{eq:norm_n} \eqref{eq:norm_d}\}
    \State scores $\gets$ TSM($N_{norm}$, $D_{norm}$) \{Eq. \eqref{eq:fitness}\}
    \State p\_r $\gets$ $C[argmax($scores$)]$ \{Target Point\}
    \State p\_last $\gets$ p\_r \{Update\}
\EndWhile
\end{algorithmic}
\caption{Feature Focus Algorithm}
\end{algorithm}

\subsubsection{Feature Focus Block}\label{subsec:active}
% \subsection{Active Lighting Approach}

The goal of the active control strategy we devise is to effectively respond to changing environmental conditions, thereby obtaining more useful 
information relevant to the given task. In the considered application, the intuition is to identify texture-rich areas, and actively illuminate them while the robot navigates. More specifically, to address the first part we propose a pipeline which consists of two main blocks: the Feature Extractor Module (FEM) and the Spot Reference Selector (SRS).

% \alberto{Driven by the aforementioned intuition, we design an active strategy specifically tailored for controlling the light source toward feature-reach areas. More specifically, the pipeline consists of three blocks: the first block handles feature extraction (FEM), the second one groups the features into suitable centroids (KG), and the final block selects the optimal centroid to be illuminated (SRS).}

The FEM module processes the enhanced image provided by the ENB module with a feature extractor algorithm in order to produce a list of key points at pixel level $[(x_1,y_1), \dots, (x_n, y_n)]$ corresponding to the features coordinates in the image. 
%It is important to note that, for consistency, the choice of the feature extractor algorithm is fixed by the V-SLAM or VO strategy to be used. 
Once the key points are extracted, they are fed into a clustering algorithm, which extracts suitable clusters and returns the coordinates of the respective centroids along with the number of features each centroid represents. Lastly, the Spot Reference Selector processes the centroids and outputs a single target point $p^r=(x^r, y^r)$ in the image plane. For the target point selection, directly using the centroid with the most number of features might seem a reasonable choice since, by problem definition, we are interested in lighting up the information-richest area. However, in low-light conditions, the noise in the image can cause fluctuations in the detected features, and the target point could frequently change in situations where multiple clusters have a similar number of features, leading to the light source continuously moving from one cluster to another. Hence, we develop a novel \textit{Target Selection Metric} (TSM) $m_i(n, d): \mathbb{R}^2 \rightarrow \mathbb{R}$ for the target point identification. In particular, it takes into account the number of features of the various centroids and their respective distance to the previously selected target point $p_{i-1}^r=(x_{i-1}^r, y_{i-1}^r)$. The TSM is defined as:
\begin{equation}
    \label{eq:fitness}
    m_i(n, d)= \alpha \cdot \frac{1}{d'_{i} + \epsilon} + (1 - \alpha) \cdot n'_{i}
\end{equation}
\alberto{where $\epsilon$ prevents the denominator from becoming zero, $\alpha$ is a tuning hyper-parameter that balances the two terms of the metric, and $n'_{i}$ represents the normalized number of features for the $i$-th centroid}
\begin{equation}
    \label{eq:norm_n}
    n'_{i} = \frac{n_i - \max({N})}{\max({N}) - \min({N})}
\end{equation}
\alberto{with $N=\{n_0, \dots, n_i, \dots, n_n\}$, and $d'_{i}$ represents the normalized distance of the $i$-th centroid w.r.t. the previously selected target point}
\begin{equation}
    \label{eq:norm_d}
    d'_{i} = \frac{d_i - \max({D})}{\max({D}) - \min({D})}
\end{equation}
\alberto{with $D=\{d_0, \dots, d_i, \dots, d_n\}$ and $d_i = || p_{i-1}^r - p_i^r ||$.}

\alberto{Lastly, as detailed in Algorithm \ref{algo:active}, the TSM is used to calculate a score for each centroid, and the centroid with the highest value is chosen as the target point to be illuminated.}

\begin{table*}[t]
\renewcommand{\arraystretch}{1.3}
\centering
\caption{Experimental results comparing our approach against the baselines in different scenarios \vspace{-1.0em}}
\resizebox{2.0\columnwidth}{!}{
\label{tab:results}
% \begin{tabular}{c|>{\centering\arraybackslash}p{1.6cm}|>{\centering\arraybackslash}p{1.6cm}|>{\centering\arraybackslash}p{1.6cm}|>{\centering\arraybackslash}p{1.6cm}|>{\centering\arraybackslash}p{1.6cm}|>{\centering\arraybackslash}p{1.6cm}|>{\centering\arraybackslash}p{1.6cm}|>{\centering\arraybackslash}p{1.6cm}|>{\centering\arraybackslash}p{1.6cm}|>{\centering\arraybackslash}p{1.6cm}|>{\centering\arraybackslash}p{1.6cm}|>{\centering\arraybackslash}p{1.6cm}|>{\centering\arraybackslash}p{1.6cm}|>{\centering\arraybackslash}p{1.6cm}|>{\centering\arraybackslash}p{1.6cm}|>{\centering\arraybackslash}p{1.6cm}}
\begin{tabular}{c|c|c|c|c|c|c|c|c|c|c|c|c|c|c|c|c}
\hline
\hline
\multirow{3}{*}{Method} & \multicolumn{16}{c}{Experimental Scenarios and Metrics}\\
\cline{2-17}
& \multicolumn{4}{c|}{Corridor\_L1} & \multicolumn{4}{c|}{Corridor\_L2} & \multicolumn{4}{c|}{Corridor\_R} & \multicolumn{4}{c}{Room\_R}\\
\cline{2-17}
& $\text{ATE}$ [$\SI{}{\meter}$] & $\text{ARE}$ [$\SI{}{rad}$] & $R_{traj}$ & $T_{lost}$ & $\text{ATE}$ [$\SI{}{\meter}$] & $\text{ARE}$ [$\SI{}{rad}$] & $R_{traj}$ & $T_{lost}$ & $\text{ATE}$ [$\SI{}{\meter}$] & $\text{ARE}$ [$\SI{}{rad}$] & $R_{traj}$ & $T_{lost}$ & $\text{ATE}$ [$\SI{}{\meter}$] & $\text{ARE}$ [$\SI{}{rad}$] & $R_{traj}$ & $T_{lost}$ \\
\hline
AL-VO (Our) & $ \cellcolor{green!40} 0.12$ & $ \cellcolor{green!40} 0.51$ & $ \cellcolor{green!40} 1.0$ & $ \cellcolor{green!40} 0$ & $ \cellcolor{green!40} 0.29$ & $ \cellcolor{red!40} 0.65$ & $ \cellcolor{green!40} 1.0$ & $ \cellcolor{green!40} 0$ & $ \cellcolor{green!40} 0.76$ & $ \cellcolor{green!40} 0.38$ & $ \cellcolor{green!40} 0.92$ & $ \cellcolor{green!40} 0$ & $ \cellcolor{green!40} 0.26$  & $ \cellcolor{red!40} 0.44$ & $ \cellcolor{green!40} 1.0$ & $ \cellcolor{green!40} 0$\\
\hline
FL-VO & $ \cellcolor{red!30} 0.37$ & $ \cellcolor{red!30} 2.54$ & $ \cellcolor{red!30} 0.37$ & $ \cellcolor{red!30} 2$ & $ \cellcolor{red!30} 1.04$ & $ \cellcolor{green!30} 0.56$ & $ \cellcolor{red!30} 0.10$ & $ \cellcolor{red!30} 1$ & $ \cellcolor{red!30} 0.91$ & $ \cellcolor{red!30} 1.12$ & $ \cellcolor{red!30} 0.18$ & $ \cellcolor{red!30} 1$ & $ \cellcolor{red!30} 1.05$ & $ \cellcolor{green!30} 0.32$ & $ \cellcolor{red!30} 0.96$ & $ \cellcolor{green!30} 0$ \\
\hline
\hline
LO-VO & $ \cellcolor{white!30} 0.02$ & $ \cellcolor{white!30} 0.79$ & $ \cellcolor{white!30} 1.0$ & $ \cellcolor{white!30} 0$ & $ \cellcolor{white!30} 0.04$ & $ \cellcolor{white!30} 0.11$ & $ \cellcolor{white!30} 1.0$ & $ \cellcolor{white!30} 0$ & $ \cellcolor{white!30} 0.05$ & $ \cellcolor{white!30} 0.14$ & $ \cellcolor{white!30} 1.0$ & $ \cellcolor{white!30} 0$ & $ \cellcolor{white!30} 0.17$  & $ \cellcolor{white!30} 0.09$ & $ \cellcolor{white!30} 1.0$ & $ \cellcolor{white!30} 0$\\
\hline
EG-VO \cite{jiang2021enlightengan} & $ \cellcolor{white!40} 0.02$ & $ \cellcolor{white!40} 0.81$ & $ \cellcolor{white!40} 1.0$ & $ \cellcolor{white!40} 0$ & $ \cellcolor{white!40} 0.09$ & $ \cellcolor{white!40} 0.22$ & $ \cellcolor{white!40} 1.0$ & $ \cellcolor{white!40} 0$ & $ \cellcolor{white!40} 0.05$ & $ \cellcolor{white!40} 0.64$ & $ \cellcolor{white!40} 0.81$ & $ \cellcolor{white!40} 0$ & $ \cellcolor{white!40} 0.16$  & $ \cellcolor{white!40} 0.08$ & $ \cellcolor{white!40} 1.0$ & $ \cellcolor{white!40} 0$\\
\hline
\hline
\end{tabular}}
% \vspace{-1em}
\end{table*} 

% \francesco{As depicted in \figref{fig:framework_overview}, each image at a low rate processed by the enhancing network is further analyzed from two main sub-blocks, the Feature Extractor Module (FEM) and the Spot Reference Generator (SRG). The FEM module uses an extractor demanded to generate meaningless features in the enhanced image. In this work, we used ORB features \cite{rublee2011orb} to produce a list of key points at pixel level $[(x1,y1), \dots (x_n, y_n)]$. The Spot Reference Generator then processes the key points that output a single target point in the image plane $T_P=(x_t,y_t)$. The target point is computed involving a clustering process, in particular, DBSCAN \cite{ester1996density} that tries to identify areas within the image that contain an adequate number of points. The concept of adequate and the hyperparameters tuning procedure is widely covered in the next section \ref{subsec:active}. Once the $T_P$ is calculated, an arm controller job is demanded for mapping the pixel level $(x,y)$ coordinates into set point values for the 2DOF arm: aiming the light to the desired point.}

 \begin{figure}[t]
     \centering
     \includegraphics[width=\linewidth]{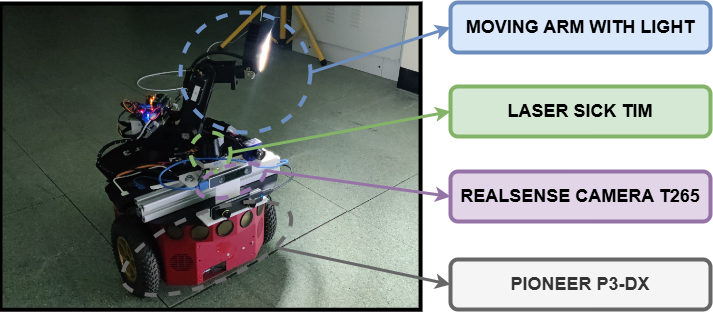}
     \caption{The experimental setup centers on a P3-DX differential mobile robotic platform (gray). This platform is equipped with a RealSense T265 camera (purple), operating in monocular mode to run the VO pipeline. The 2DLiDAR system (green) is demanded to guarantee safety (i.e., emergency obstacle avoidance) and ground truth (GT) pose generation. On the top of the chassis (blue) is the moving arm with attached the lighting device used to light up the surrounding environment.}
     \label{fig:robot_overview}
     \vspace{-1em}
\end{figure}

\subsubsection{Arm Controller Block} \label{subsec:arm_controller}
Given the position of the target point $p^r$, the arm controller block converts the pixel coordinates into a target position for the 2DOF arm. Two joints, $J_1$, and $J_2$, control the base rotation and light tilt, respectively. Servo motors govern joint positions, accepting Pulse-Position Modulation (PPM) values as input, internally mapped to specific rotor angles.
%Our low-level controller relies on the Robot Operating System (ROS) and is designed to interface with the main computer via a serial port. 
Upon receiving a desired position message, the controller applies a smoothing function with saturation within a predefined range to limit maximum acceleration. The image corners are mapped to the corresponding PPM values defining: (i) the upper and lower bounds and (ii) a linear function for converting $x$ and $y$ pixel coordinates to joint positions $f(x,y)\rightarrow(\text{PPM}_{J_1}, \text{PPM}_{J_2})$. In general, linear coordinates mappings employing a wide-angle camera can lead to inaccuracies due to radial distortions, especially near the edges of the image. However, the light source produces a beam of $30^{\circ}$, so the error between the centroid and the aiming can be considered negligible. The modularity of the system allows changing the mapping function without redesigning the entire system. 

\section{EXPERIMENTS}\label{sec:experiments}

\subsection{Experimental Setup}\label{subsec:exp_setup}
The robot involved in the experimental session is a differential P3-DX robotic platform equipped with multiple devices, including  2D sick LiDAR, an intel T265 camera, and an onboard Nvidia Jetson TX1. 
% \textcolor{red}{The embedded computer enables the usage of Linux and ROS, integrating all peripherals into distinct ROS nodes, including mission planning and lighting control.} 
The sick TIM551 Lidar is used for creating the map, and the GT poses for the performance evaluation. The T265 camera module was set to acquire $848\times800$ grayscale images at $\approx$ $30$ FPS with a fixed exposure time of $16 ms$. The robotic arm, a 2DOF, with the two joints, $J_1$ and $J_2$ controlled by two independent PPM signals, acts as pan-tilt support with a $\SI{30}{\watt}$ light source end effector. The lighting device has a narrow and fixed beam angle $\approx 30^{\circ}$ that guarantees in-depth illumination. In \figref{fig:robot_overview}, we report an overview of the robotic platform. 

The VO pipeline adopted for the experimental campaign is ORB-SLAM3\cite{ORBSLAM3_TRO}, configured with its original parameters. Moreover, for consistency, we selected ORB as the feature extractor in the \textit{Feature Focus Block} implementation.

\begin{figure*}[ht]
    \centering
    \begin{minipage}[c]{0.99\linewidth}

    \raisebox{0.7 cm}{\rotatebox[origin=l]{90}{\makebox[1cm]{AL-VO}}}\hspace{0.1cm}
    \subfigure{\includegraphics[width=0.191\linewidth, height=0.16\linewidth]{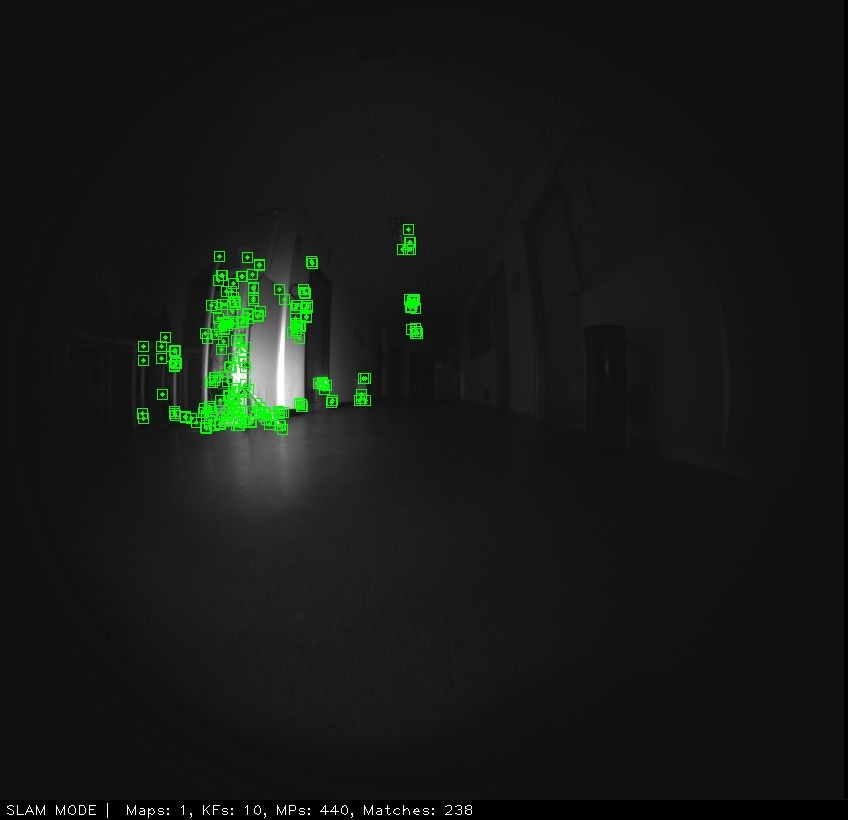}}
    \subfigure{\includegraphics[width=0.191\linewidth, height=0.16\linewidth]{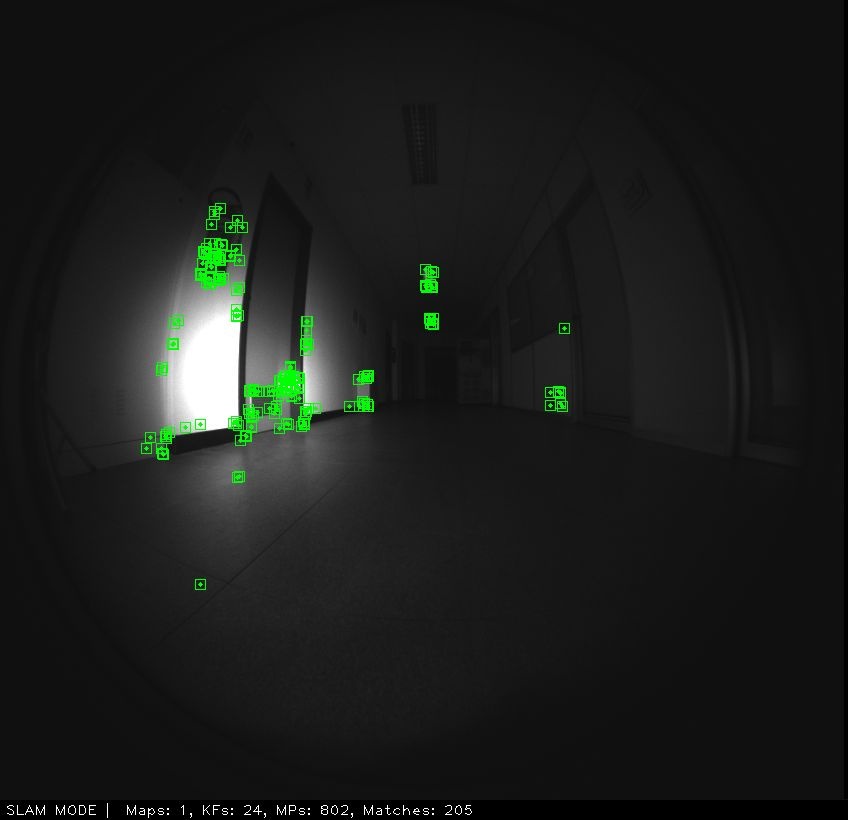}} 
    \subfigure{\includegraphics[width=0.191\linewidth, height=0.16\linewidth]{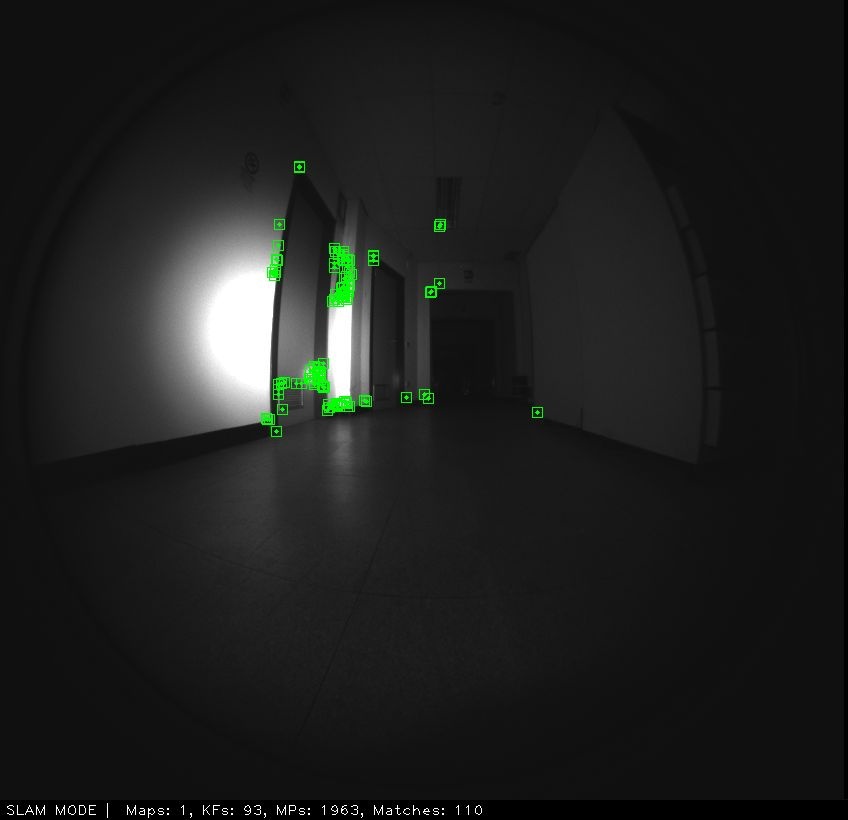}}
    \subfigure{\includegraphics[width=0.191\linewidth, height=0.16\linewidth]{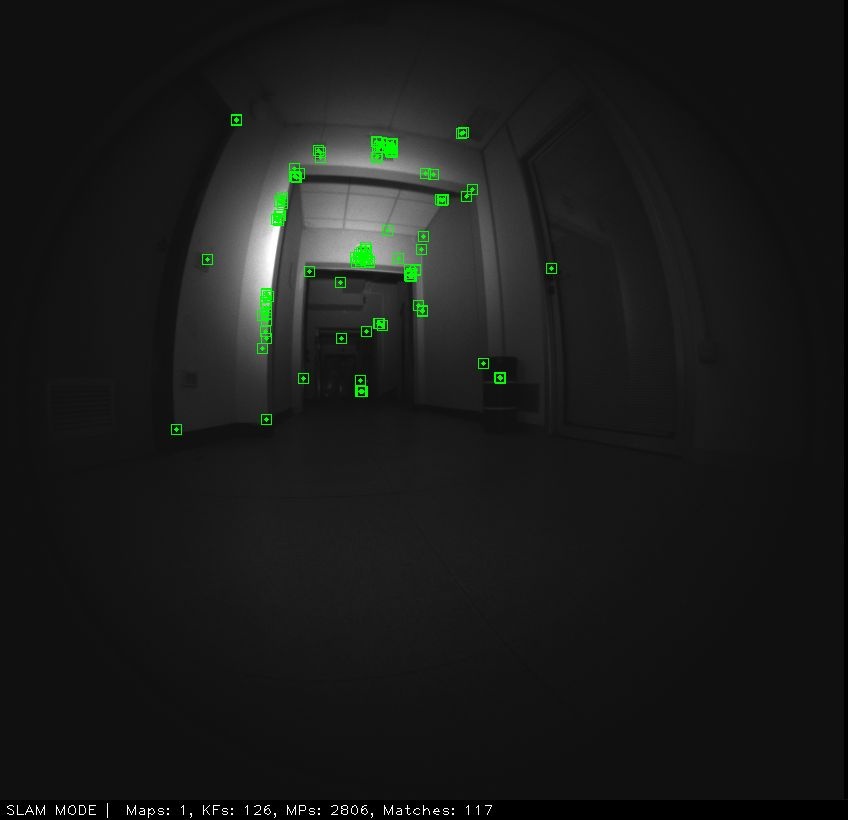}}
    \subfigure{\includegraphics[width=0.191\linewidth, height=0.16\linewidth]{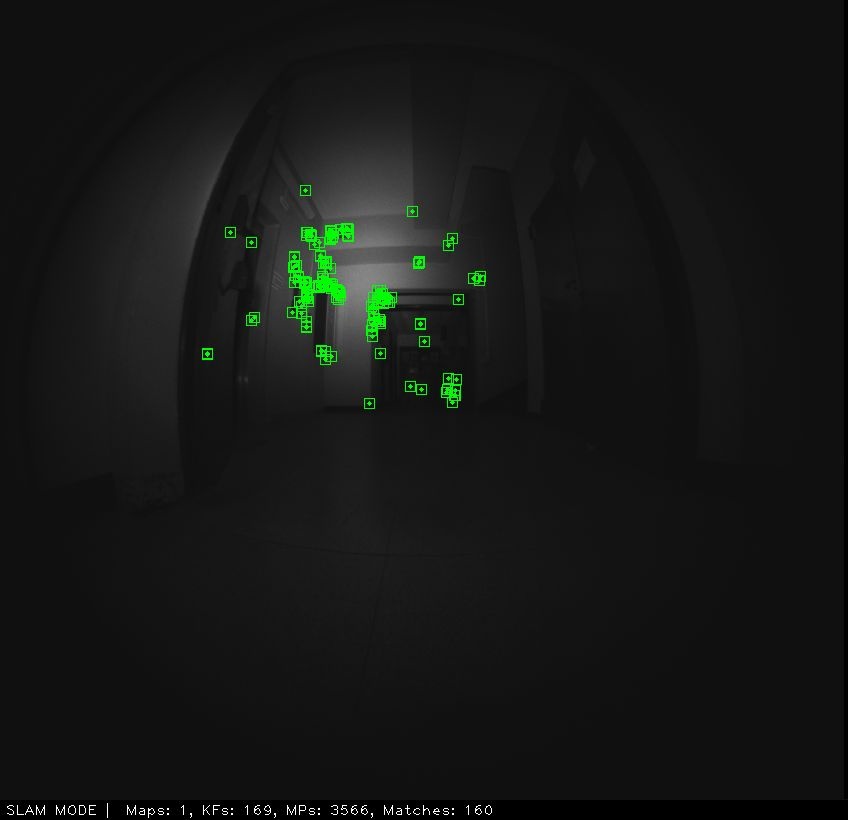}}

    \raisebox{0.7 cm}{\rotatebox[origin=l]{90}{\makebox[1cm]{FL-VO}}}\hspace{0.1cm}
    \subfigure{\includegraphics[width=0.191\linewidth, height=0.16\linewidth]{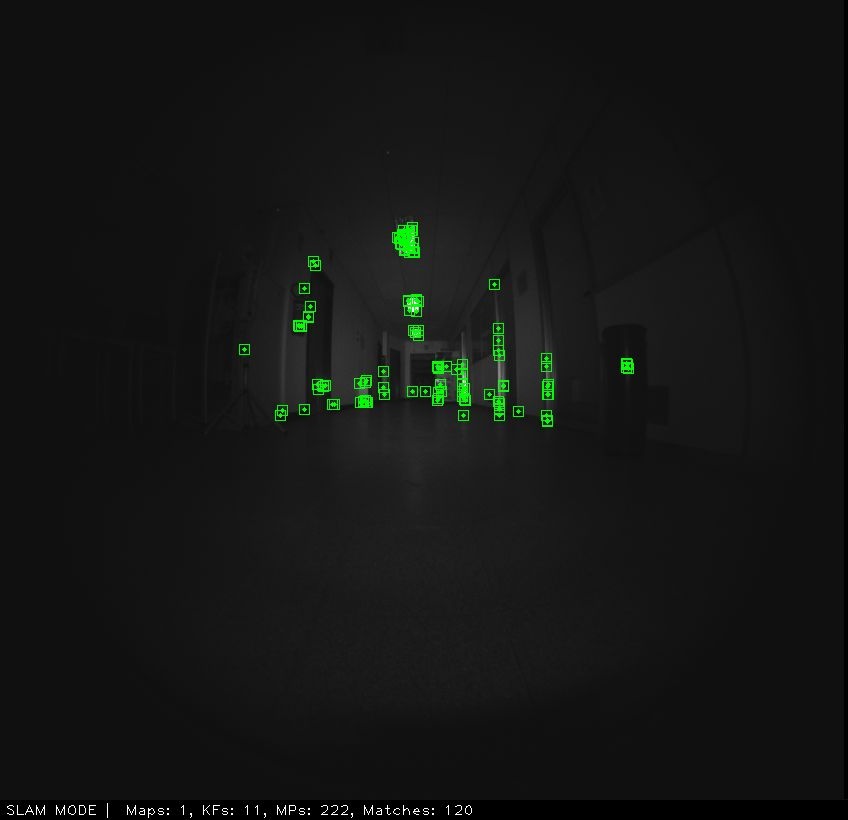}}
    \subfigure{\includegraphics[width=0.191\linewidth, height=0.16\linewidth]{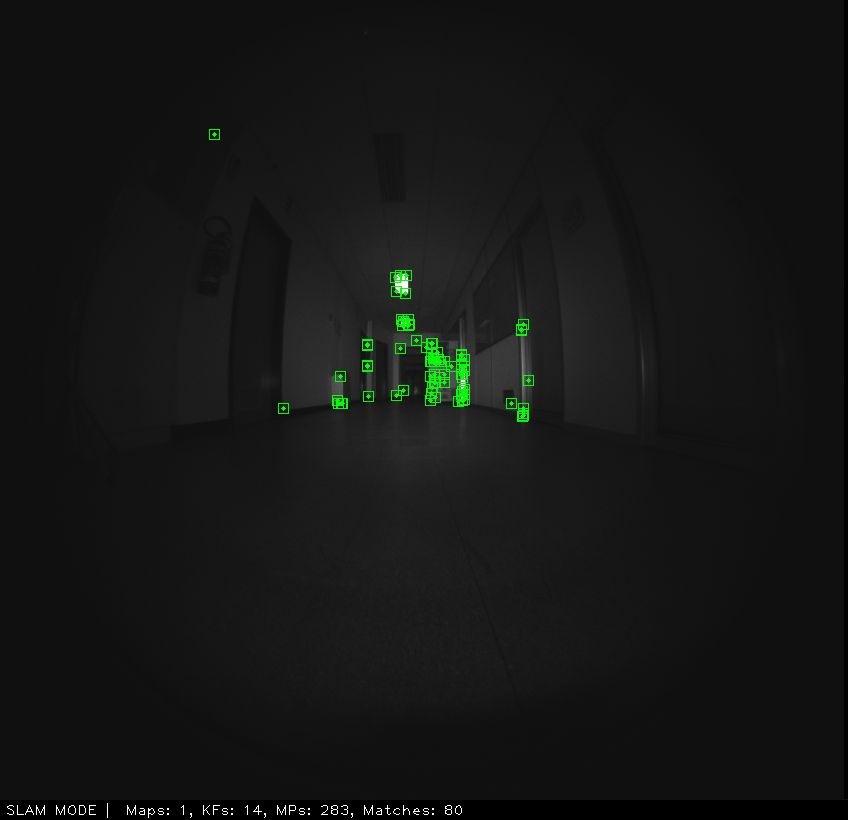}} 
    \subfigure{\includegraphics[width=0.191\linewidth, height=0.16\linewidth]{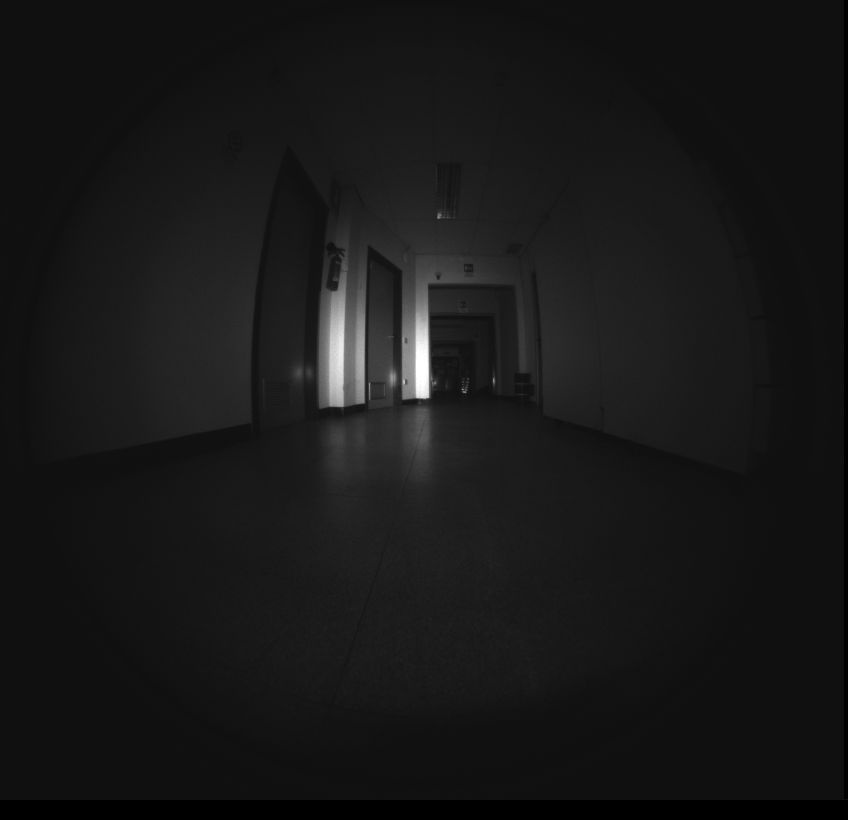}}
    \subfigure{\includegraphics[width=0.191\linewidth, height=0.16\linewidth]{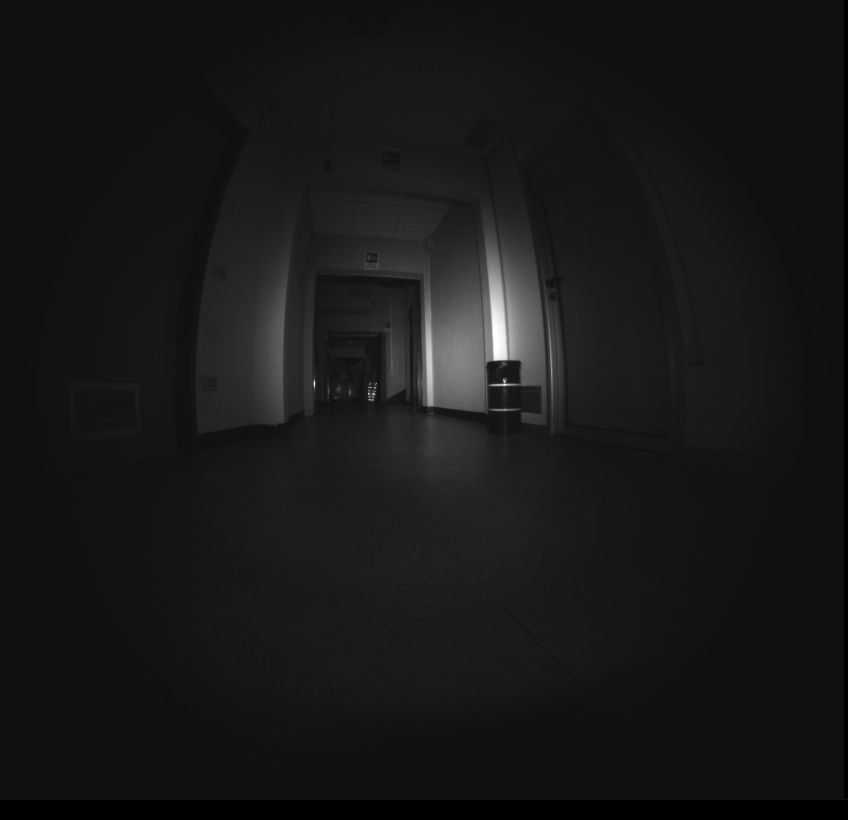}}
    \subfigure{\includegraphics[width=0.191\linewidth, height=0.16\linewidth]{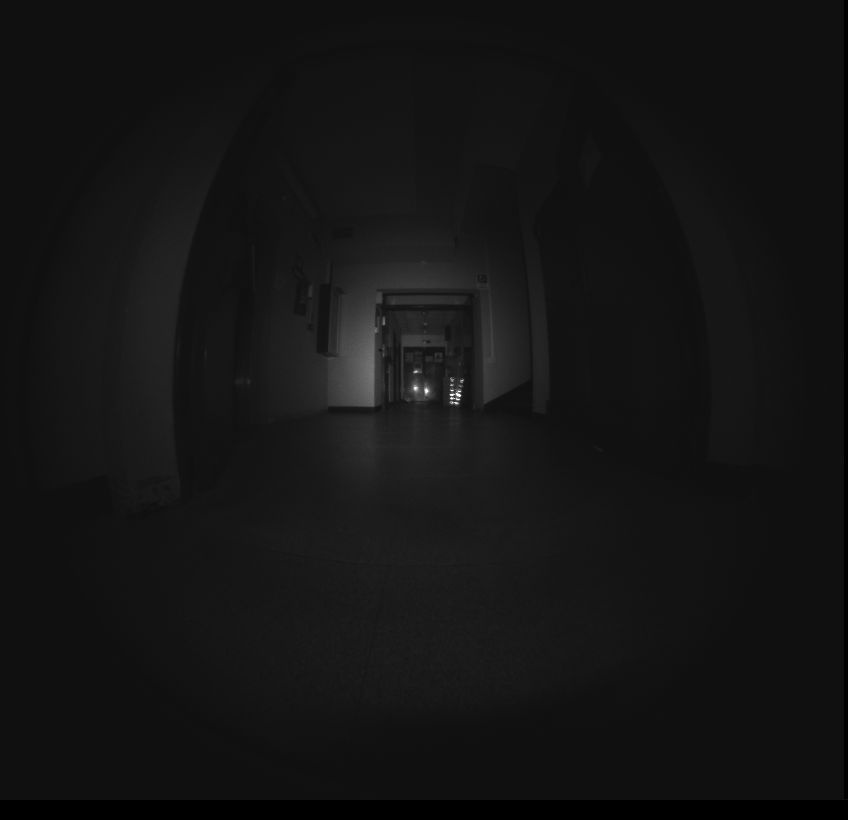}}
    
    \raisebox{0.7 cm}{\rotatebox[origin=l]{90}{\makebox[1cm]{LO-VO}}}\hspace{0.1cm}
    \subfigure{\includegraphics[width=0.191\linewidth, height=0.16\linewidth]{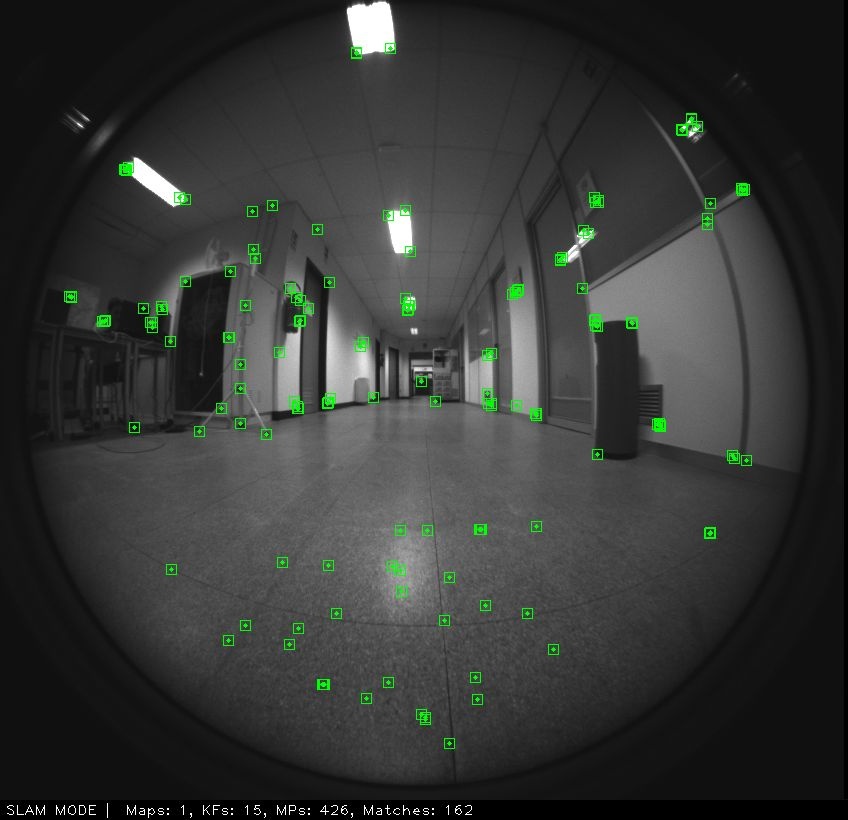}}
    \subfigure{\includegraphics[width=0.191\linewidth, height=0.16\linewidth]{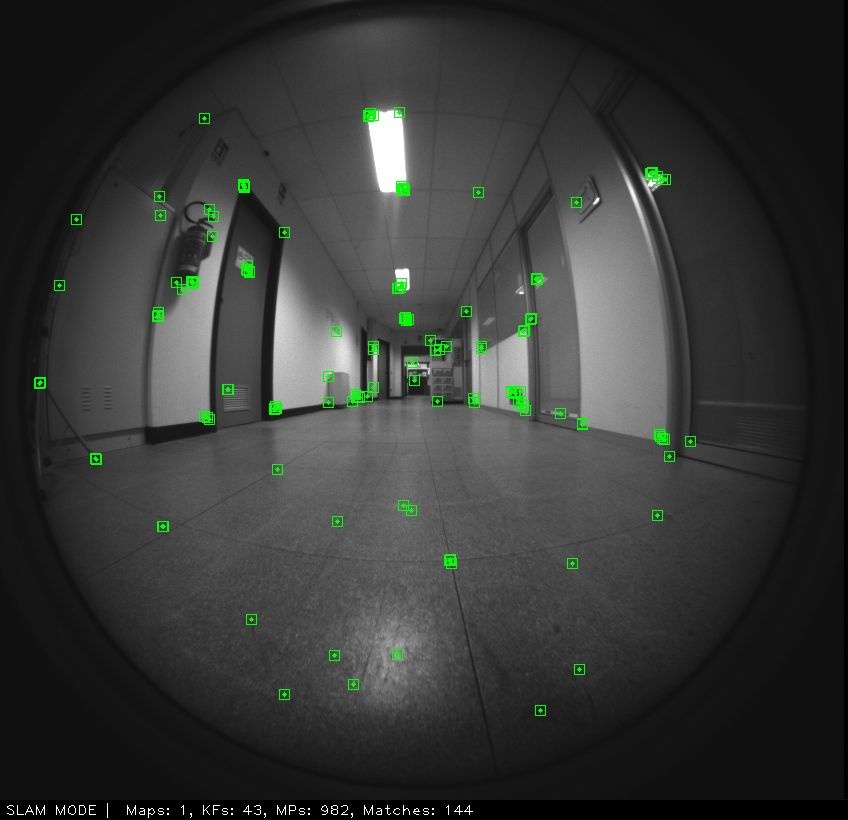}} 
    \subfigure{\includegraphics[width=0.191\linewidth, height=0.16\linewidth]{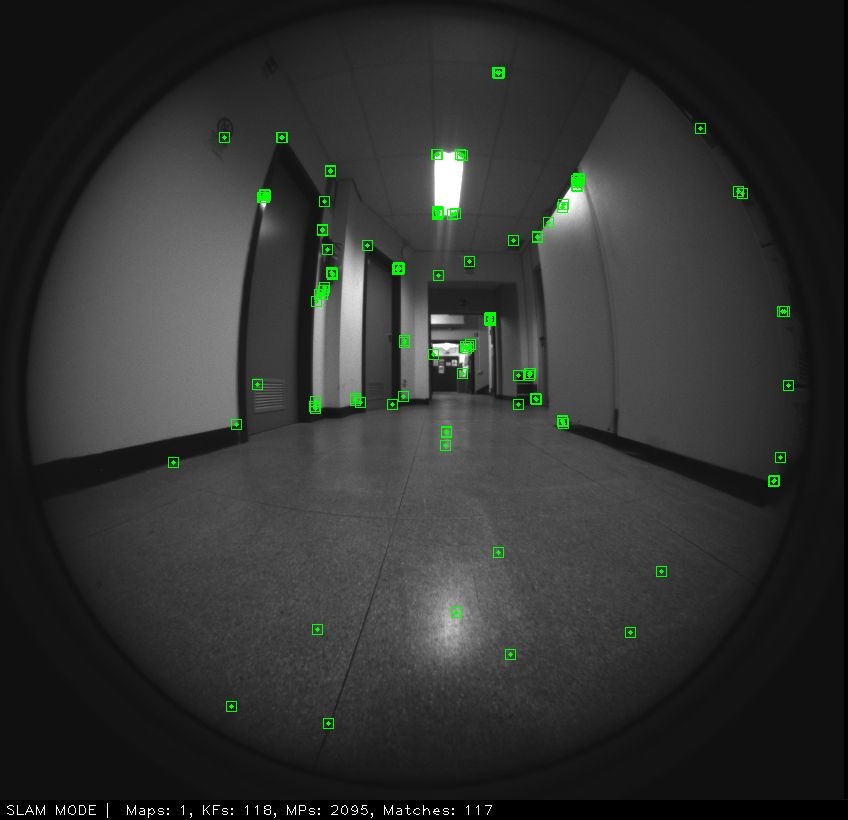}}
    \subfigure{\includegraphics[width=0.191\linewidth, height=0.16\linewidth]{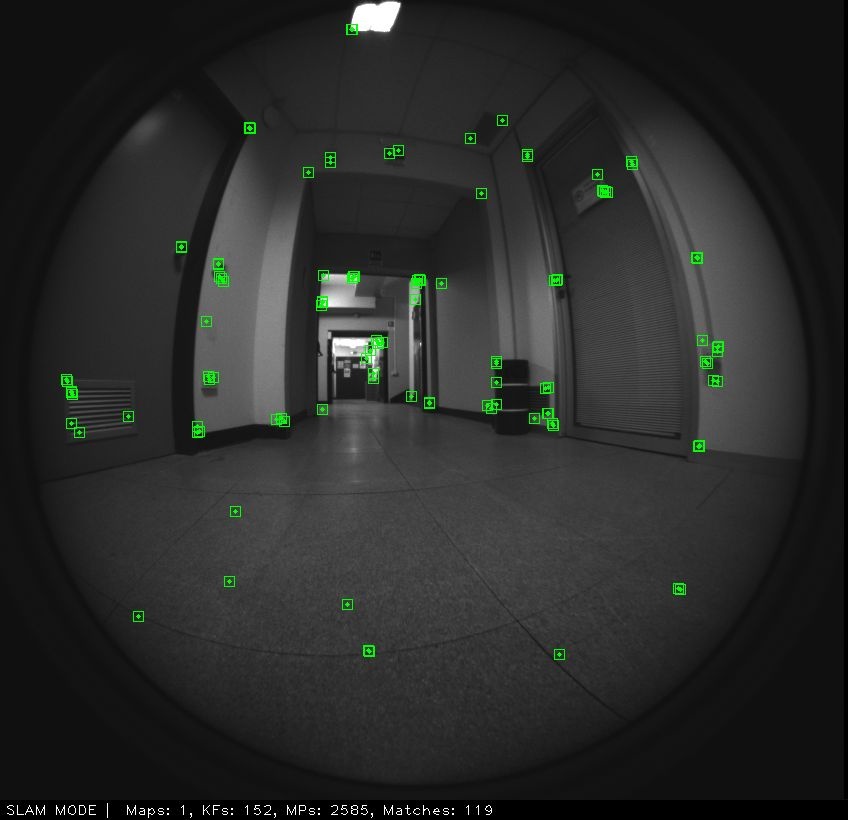}}
    \subfigure{\includegraphics[width=0.191\linewidth, height=0.16\linewidth]{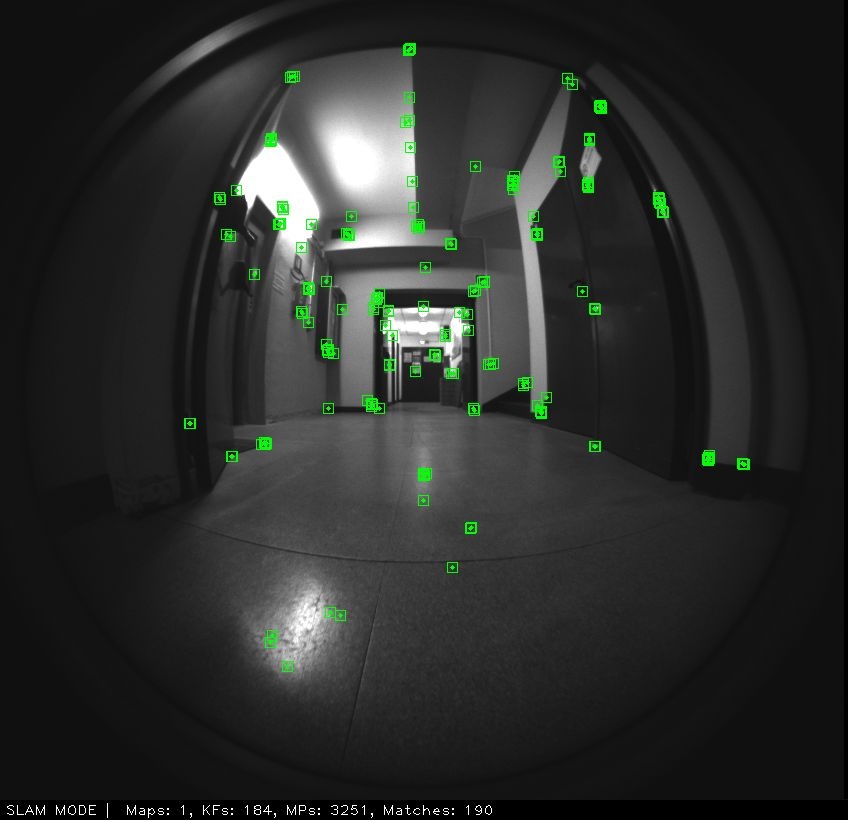}}

    \end{minipage}
    
    \caption{\enspace Snapshots of the corridor experiment. By dynamically adjusting the light beam, \textbf{AL-VO} successfully tracks features and completes the trajectory, while the fixed illumination method \textbf{FL-VO} struggles in low-texture areas. \textbf{LO-VO} represents the optimal performance, obtainable only under ideal lighting.\vspace{-2em}}   \label{fig:exp_setup}
\end{figure*}

\subsection{Experimental Scenario and Baselines}
Shadows, direct light, and reflections can degrade visual data quality, while environments lacking texture offer few features for tracking, hindering accurate image matching. These challenges worsen in darkness.
To evaluate our proposed method, we selected four indoor environments considering texture richness, obstacles, and path complexity. 
The first two scenarios are rich in textures (denoted as R) and are located in a basement space comprising different rooms, elevators, and a central corridor. The first one, \textit{Corridor\_R}, represents a straight corridor of $\SI{30}{\meter}$. The second one, \textit{Room\_R} involves navigating a room with two doors, one used as an entrance and the other as an exit. The start and the end points are very close together. The last two scenarios, \textit{Corridor\_L1} and \textit{Corridor\_L2} are wide corridors characterized by a low amount of texture located mainly on the sides of the scene. 
These last two scenarios allows to assess whether moving light to areas identified by the FEM block results in an effective benefit for the VO pipeline. 

Our method is referred to as \textbf{AL-VO} (\textbf{A}ctive \textbf{L}ight \textbf{V}isual \textbf{O}dometry) and, as comparison baselines, we use the following methodologies:
\begin{itemize}
    \item \textbf{FL-VO} (\textbf{F}ixed \textbf{L}ight \textbf{V}isual \textbf{O}dometry): the robotic arm is held at the center position, e.g., a fixed centroid $C(W/2, H/2)$, where $W$ and $H$ are the Width and Height of the image.
    \item \textbf{LO-VO} (\textbf{L}ights \textbf{O}n \textbf{V}isual \textbf{O}dometry): the scenario is lit by the light of the building and during daylight.
    \item \textbf{EG-VO} (\textbf{E}nlighten\textbf{G}an \textbf{V}isual \textbf{O}dometry): same environmental condition of FL-VO, but the images are enhanced with \cite{jiang2021enlightengan}. 
\end{itemize}

\subsection{Metrics}
To assess the effectiveness of the proposed AL-VO method, we evaluated the experimental results using three metrics computed by the EVO toolbox \cite{grupp2017evo}. In particular, we rely on the Absolute Pose Error (APE), which comprises translational and rotational components. Defining the Ground Truth (GT) and estimated poses the at timestamp $n$ as $\overset{*}{P_{n}},\overset{-}{P_{n}} \in SE(3)$ respectively, the APE error $E_n $ can be defined as:
\begin{equation}\label{eq:err_n}
    E_n = {\overset{*}{P_{n}}} \ominus \overset{-}{P_{n}} \in SE(3) ,
\end{equation}
where the inverse composition operator $\ominus$ takes two poses and gives the relative pose  \cite{lu1997robot}.
The error $E_n$ can be decomposed into the Absolute Translation Error (ATE) and the Absolute Rotational Error (ARE) defined as follows:
\begin{align}
\text{ATE}_{n} &= \lVert \text{trans}(E_n) \rVert \label{eq:ate}\\
\text{ARE}_{n} &= \lvert \text{angle}(\log_{SO(3)}(\text{rot}(E_n))) \rvert \label{eq:are},
\end{align}
where $log_{SO(3)}(\cdot)$ is the inverse of $exp_{SO(3)}(\cdot)$.
In the following, we refer to ATE and ARE as the Root Mean Squared Error (RMSE) of \equatref{eq:ate} and \equatref{eq:are}, respectively.

In addition to the RMSE values, we also consider the ratio $R_{t}$ defined as:
\begin{equation}
R_{t} = \nicefrac{\overset{-}{L}}{\overset{*}{L}} \quad , \; R_{t} \in \{\:\mathbb{R}\; | \; R_{t} \geq 0 \:\} \; ,
\label{eq:ratio_traj}
\end{equation}
where $\overset{*}{L}$ and $\overset{-}{L}$ denote the lengths of ground truth and estimated trajectories. 
The reference value is $R_{t} = 1$, meaning that the estimated and GT trajectories have the same length. Values below or above this value suggest the presence of accumulated errors in the estimation process. 
The more the values diverge from the reference, the higher the chances of critical failures, including losing track of its position entirely (e.g., due to a sudden and drastic environmental change).
We included the $R_{t}$ value to avoid erroneous conclusions from just RMSE scores: an incomplete but highly accurate trajectory would yield lower values than a whole trajectory with a higher mean error. Moreover, as an additional metric in \tabref{tab:results}, we included a special counter $T_{lost}$ that counts the number of tracking lost during the pose estimation process. 

\begin{figure}[t]
    \centering
    \begin{minipage}[c]{0.99\linewidth}
    \centering 

    \subfigure[]{\includegraphics[width=.49\linewidth]{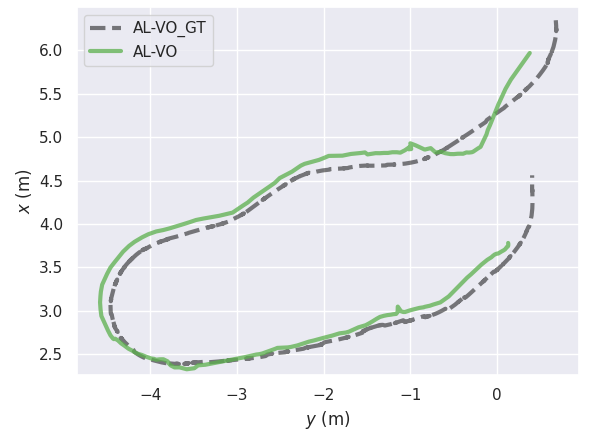}\label{fig:AL_VO}}
    \subfigure[]{\includegraphics[width=.49\linewidth]{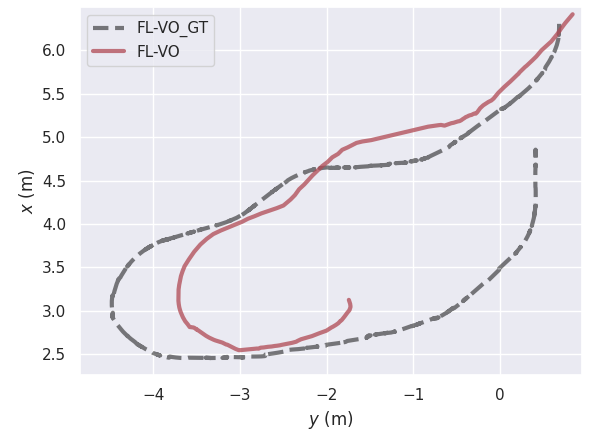}\label{fig:FL_VO}}
    \subfigure[]{\includegraphics[width=.49\linewidth]{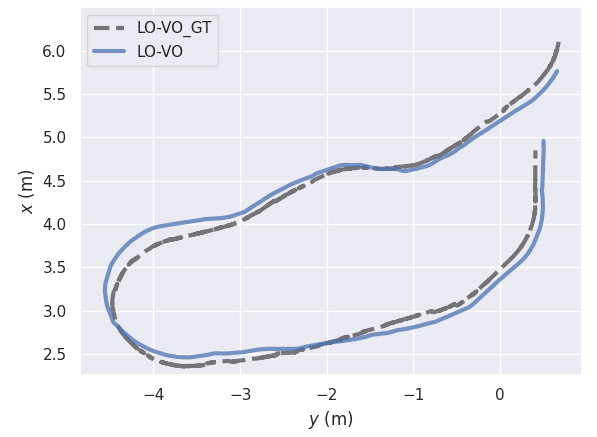}\label{fig:LO-VO}}
    \subfigure[]{\includegraphics[width=.49\linewidth]{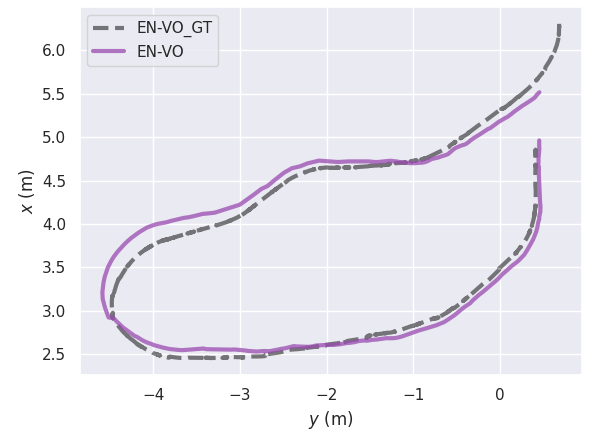}\label{fig:EN-VO}}

    \end{minipage}
    \caption{\enspace Plot of the (a) \textbf{AL-VO} (b) \textbf{FL-VO} (c) \textbf{LO-VO} and (d) \textbf{EN-VO} trajectories w.r.t. the Ground Truth in the \textit{Room\_R} scenario. \vspace{-1.5em}}
    \label{fig:dataset_samples}
\end{figure}

\subsection{Results}

The results of the experimental campaign are reported in \tabref{tab:results}. The first important finding is that the active approach \textbf{AL-VO} outperforms the fixed light counterpart \textbf{FL-VO} in almost all scenarios with respect to the defined metrics. More specifically, in challenging scenarios like the \textit{Corridor} environments,  \textbf{FL-VO} constantly fails and cannot finish the trajectory. As shown in \figref{fig:exp_setup}, this is caused by the flat surfaces that \textbf{FL-VO} accidentally illuminates during the navigation, providing insufficient texture for the VO algorithm to work properly. On the other hand, thanks to the developed active strategy, \textbf{AL-VO} can redirect the light beam towards areas rich with corners, edges, and objects, allowing \textbf{AL-VO} to complete all the trajectories without losing tracking ($T_{lost}=0$ and $R_t\approx1$). For reference purposes only, we have also included the performance metrics obtained by the VO pipeline under optimal light conditions (\textbf{LO-VO}) and in low-light conditions using the EnlightenGAN (\textbf{EG-VO}). As expected, \textbf{LO-VO} and \textbf{EG-VO} achieve the best overall results. Furthermore, it should be noticed that \textbf{EG-VO} obtains remarkable performance in dark conditions, achieving results very close to those obtained by \textbf{LO-VO} (daylight). It is important to highlight that while EnlightenGAN offers excellent enhancement, its computational complexity hinders the deployment of \textbf{EG-VO} on resource-constrained robotic platforms. Therefore, \textbf{AL-VO} represents a better choice for low-light VO in such scenarios. %\textcolor{green}{The computational complexity of EnlightenGAN, which enables its superior enhancement performance, limits the deployment of \textbf{EG-VO} on robotic platforms with limited resources.  For this reason, \textbf{AL-VO} represent a more reasonable algorithm for low-light VO for such platforms.} %\textcolor{red}{The computational cost of the EnlightenGAN makes it unfeasible to deploy \textbf{EG-VO} on a robotic platform with limited resources, which makes \textbf{AL-VO} a better choice to perform VO in low-light scenarios.}
Lastly, as also shown by the qualitative results reported in \figref{fig:dataset_samples}, every method was able to produce a reasonable trajectory in the \textit{Room\_R} scenario. However, while \textbf{FL-VO} exhibits a marked deviation from the GT in one of the curved sections, \textbf{AL-VO} is able to maintain a much closer alignment w.r.t the GT trajectory, and manages to achieve scores very close to that obtained by \textbf{LO-VO} and \textbf{EN-VO}.

\section{CONCLUSION}\label{sec:conclusion}
In this work, we introduced a novel active lightning framework designed to enable VO and V-SLAM algorithms to properly operate in dark scenarios. 
Specifically, we proposed a new active method to identify and illuminate the portion of the image with the highest number of features.
Extensive real-world experiments validated the effectiveness of our approach and highlighted its potential for practical deployment.
In future work, we will enhance our method by incorporating in the \textit{Target Selection Metric} an additional term to take into account the future movements of the robot.


%%%%%%%%%%%%%%%%%%%%%%%%%%%%%%%%%%%%%%%%%%%%%%%%%%%%%%%%%%%%%%%%%%%%%%%%%%%%%%%%

\bibliographystyle{IEEEtran}
\bibliography{bibliography}

\end{document}